\documentclass[10pt,twocolumn,letterpaper]{article}
\usepackage{cvpr}
\usepackage{times}
\usepackage{helvet}
\usepackage{courier}
\usepackage{url}
\usepackage{graphicx}
\usepackage{amsmath}
\usepackage{algorithm}
\usepackage[pagebackref=true,breaklinks=true,letterpaper=true,colorlinks,bookmarks=false]{hyperref}
\usepackage{cleveref}
\usepackage{booktabs}
\usepackage{amssymb}
\usepackage{stfloats}
\usepackage{graphics}
\usepackage{subcaption}
\usepackage[noend]{algpseudocode}
\DeclareMathOperator{\sign}{sign}
\DeclareMathOperator{\grad}{grad}

\cvprfinalcopy
\author{Valentin Khrulkov\\
Skolkovo Institute of Science and Technology\\
143025, Nobel St. 3, Skolkovo Innovation Center \\
Moscow, Russia\\
{\tt\small valentin.khrulkov@skolkovotech.ru}
\and
Ivan Oseledets\\
Skolkovo Institute of Science and Technology\\
Institute of Numerical Mathematics of the\\
Russian Academy of Sciences \\
119333, Gubkina St. 8 \\
Moscow, Russia \\
{\tt\small i.oseledets@skoltech.ru}
}

\def\cvprPaperID{3771}

\begin{document}
\title{Art of singular vectors and universal adversarial perturbations}
\maketitle
\begin{abstract}
Vulnerability of Deep Neural Networks (DNNs) to adversarial attacks has been attracting a lot of attention in recent studies. It has been shown that for many state of the art DNNs performing image classification there exist universal adversarial perturbations --- image-agnostic perturbations mere addition of which to natural images with high probability leads to their misclassification. In this work we propose a new algorithm for constructing such universal perturbations. Our approach is based on computing the so-called $(p, q)$-singular vectors of the Jacobian matrices of hidden layers of a network. Resulting perturbations present interesting visual patterns, and by using only 64 images we were able to construct universal perturbations with more than 60 \% fooling rate on the dataset consisting of 50000 images. We also investigate a correlation between the maximal singular value of the Jacobian matrix and the fooling rate of the corresponding singular vector, and show that the constructed perturbations generalize across networks.
\end{abstract}

\section{Introduction}
Deep Neural Networks (DNNs) with great success have been applied to many practical problems in computer vision \cite{lecun1995convolutional,szegedy2015going,he2016deep} and in audio and text processing \cite{graves2013speech,mikolov2011extensions,gers1999learning}. However, it was discovered that many state-of-the-art DNNs are vulnerable to adversarial attacks \cite{goodfellow2014explaining,moosavi2016universal,szegedy2013intriguing}, based on adding a \emph{perturbation} of a small magnitude to the image. Such perturbations are carefully constructed in order to lead to misclassification of the perturbed image and moreover may attempt to force a specific predicted class (targeted attacks), as opposed to just any class different from the ground truth (untargeted attacks).
Potential undesirable usage of adversarial perturbations in practical applications such as autonomous driving systems and malware detection has been studied in \cite{kurakin2016adversarial,grosse2016adversarial}. This also motivated the research on defenses against various kinds of attack strategies \cite{papernot2016distillation, goodfellow2016cleverhans}. 

In the recent work Moosavi \etal \cite{moosavi2016universal} have shown that there exist \emph{universal} adversarial perturbations --- image-agnostic perturbations that cause most natural images to be misclassified. They were constructed by iterating over a dataset and recomputing the "worst" direction in the space of images by solving an optimization problem related to geometry of the decision boundary. Universal adversarial perturbations exhibit many interesting properties such as their universality \emph{across networks}, which means that a perturbation constructed using one DNN will perform relatively well for other DNNs. 
 
We present a new algorithm for constructing universal perturbations based on solving simple optimization problems which correspond to finding the so-called $(p, q)$-singular vector of the Jacobian matrices of feature maps of a DNN. Our idea as based on the observation that since the norm of adversarial perturbations is typically very small, perturbations in the non-linear maps computed by the DNN can be reasonably well approximated by the Jacobian matrix. 
The $(p, q)$-singular vector of a matrix $A$ is defined as the solution of the following optimization problem
\begin{equation}
\label{eq:pq-def}
  \| A v \|_q \to \max, \quad \|v\|_p = 1,
\end{equation}
and if we desire $\|v\|_p = L$ instead, it is sufficient to multiply the solution of (\ref{eq:pq-def}) by $L$.
Universal adversarial perturbations are typically generated with a bound in the $\infty$-norm, which motivates the usage of such general construction. To obtain the $(p, q)$-singular vectors we use a modification of the standard power method, which is adapted to arbitrary $p$-norms. The main contributions of our paper are
\begin{itemize}
\item We propose an algorithm for generating universal adversarial perturbation, using the generalized power method for computing the $(p, q)$-singular vectors of the Jacobian matrices of the feature maps.
\item Our method is able to produce relatively good universal adversarial examples from a relatively small number of images from a dataset.
\item We investigate a correlation between the largest $(p, q)$-singular value and the fooling rate of the generated adversarial examples; this suggests that this singular value can be used as a quantitative measure of the robustness of a given neural network and can be in principle incorporated as the regularizer for the DNNs.
\item We analyze various properties of the computed adversarial perturbations such as generalization across networks and dependence of the fooling rate on the number of images used for construction of the perturbation.
\end{itemize}
\section{Problem statement}\label{sec:statement}
Suppose that we have a standard feed-forward DNN which takes a vector $x$ as 
the input, and outputs a vector of probabilities $p(x)$ for the class labels. Our goal given parameters $q \geq 1$ and $L > 0$ is to produce a vector $\varepsilon$ such that
\begin{equation}
\arg \max p(x) \neq \arg \max p(x+\varepsilon), \quad \|\varepsilon\|_q = L,
\end{equation}
for as many $x$ in a dataset as possible. Efficiency of a given universal adversarial perturbation $\varepsilon$ for the dataset $X$ of the size $N$ is called the \textit{fooling rate} and is defined as 
\begin{equation}
\frac{|\lbrace x \in X : \arg \max p(x) \neq \arg \max p(x+\varepsilon) \rbrace|}{N}.
\end{equation}
Let us denote the outputs of the $i$-th hidden layer of the network by $f_i(x)$. Then for a small vector $\varepsilon$ we have
$$f_i(x + \varepsilon) - f_i(x) \approx J_i(x) \varepsilon,$$ where 
$$J_i(x) = \frac{\partial f_i}{\partial x} \bigg\rvert_{x},$$
is the Jacobian matrix of $f_i$.
Thus, for any $q$-norm
\begin{equation}\label{eq:jac_perturb}
\|f_i(x + \varepsilon) - f_i(x) \|_q \approx \| J_i(x) \varepsilon \|_q,
\end{equation}
We can conclude that for perturbations which are small in magnitude in order to sufficiently perturb the output of a hidden layer, it is sufficient to maximize right-hand side of the \cref{eq:jac_perturb}. It seems reasonable to suggest that while propagating further in the network it will dramatically change the predicted label of $x$.

Thus to construct an adversarial perturbation for an individual image $x$ we need to solve
\begin{equation}\label{eq:max_problem}
\| J_i(x) \varepsilon \|_q \to \max, \quad \|\varepsilon\|_p = L,
\end{equation}
and due to homogeneity of the problem defined by \cref{eq:max_problem} it is sufficient to solve it for $\| \varepsilon \|_p = 1$. The solution of (\ref{eq:max_problem}) is defined up to multiplication by $-1$ and is called the $(p, q)$-singular vector of $J_i(x)$. Its computation in a general case is the well-known problem \cite{bhaskara2011approximating, boyd1974power}. In several cases e.g. $p = \infty, q = \infty$ and $p=1, q=1$, algorithms for finding the exact solution of the problem (\ref{eq:max_problem}) are known \cite{trefethen1997numerical}, and are based on finding the element of maximal absolute value in each row (column) of a matrix. However, this approach requires iterating over all elements of the matrix $A$ and thus has complexity $O(nm)$ for the matrix of size $n \times m$. Typical size of such a matrix appearing in our setting, e.g. taking VGG-19 network, output of the first pooling layer, and batch size of $64$ (usage of a batch of images is explained further in the text),  would be $9633792 \times 802816$, which requires roughly $30$ TB of memory to store and makes these algorithms completely impractical. In order to avoid these problems, we switch to \emph{iterative} methods. Instead of evaluating and storing the full matrix $A$ we use only the \emph{matvec function} of $A$, which is the function that given an input vector $v$ computes an ordinary product $Av$ without forming the full matrix $A$, and typically has $O(n)$ complexity. In many applications that deal with extremely large matrices using matvec functions is essentially mandatory.

For computing the $(p, q)$-singular vectors there exists a well-known Power Method algorithm originally developed by Boyd \cite{boyd1974power}, which we explain in the next section. We also present a modification of this method in order to construct \textit{universal} adversarial perturbations.
\section{Generalized power method}
Suppose that for some linear map $A$ we are given the matvec functions of $A$ and $A^{\top}$. Given parameter $r \geq 1$ we also define a function 
\begin{equation}
\psi_r(x) = \sign x |x|^{r-1},
\end{equation}
which applies to vectors element-wise. As usual for $r \geq 1$ we also define $r'$ such that $\frac{1}{r} + \frac{1}{r'}=1$. Then, given some initial condition $x$, one can apply the following \cref{alg:power} to obtain a solution of (\ref{eq:max_problem}).
\begin{algorithm}[!ht]
\caption{Power method for generating the $(p,q)$-singular vectors of a linear map $A$}\label{alg:power}
\begin{algorithmic}[1]
\State \textbf{Inputs}: initial condition $x$, the matvec functions of $A$ and $A^{\top}$
\State $x\gets \frac{x}{\|x\|_p}$\Comment $(p, q)$-singular vector
\State $s\gets \|Ax\|_q$\Comment $(p, q)$-singular value
\While{not converged}
\State $Sx\gets \psi_{p'}(A^{\top}\psi_q (Ax))$
\State $x\gets \frac{Sx}{\|Sx\|_p}$
\State $s\gets \|Ax\|_q$ 
\EndWhile
\State \textbf{return} $x, s$
\end{algorithmic}
\end{algorithm}
In the case $p=q=2$ it becomes the familiar power method for obtaining the largest eigenvalue and the corresponding eigenvector, applied to the matrix $A^{\top}A$.
\\
The discussion so far applies to finding an adversarial perturbation for an instance $x$. To produce \emph{universal} adversarial perturbation we would like to maximize the left-hand size of (\ref{eq:max_problem}) uniformly across all the images in the dataset $X$. For this we introduce a new optimization problem
\begin{equation}\label{eq:opt-big}
\sum_{x_j \in X} \| J_i(x_j) \varepsilon\|_q^q \to \max, \quad \|\varepsilon\|_p = L.
\end{equation}
A solution of the problem defined by \cref{eq:opt-big} uniformly perturbs the output of the $i$-th layer of the DNN, and thus can serve as the universal adversarial perturbation due to the reasons discussed in the introduction. Note that the problem given in \cref{eq:opt-big} is exactly equivalent to 
$$\|J_i \varepsilon\|_q \to \max, \quad \|\varepsilon\|_p = L, $$
where $J_i$ is the matrix obtained via stacking $J_i(x_j)$ vertically for each $x_j \in X$. To make this optimization problem tractable, we apply the same procedure to some randomly chosen subset of images (batch) $X_b \subset X$, obtaining 
\begin{equation}\label{eq:opt-small}
\sum_{x_j \in X_b} \| J_i(x_j) \varepsilon\|_q^q \to \max, \quad \|\varepsilon\|_p = L,
\end{equation}
and hypothesize that the obtained solution will be a good approximate to the exact solution of (\ref{eq:opt-big}). We present this approach in more detail in the next section.
\section{Stochastic power method}
Let us choose a fixed batch of images $X_b = \lbrace x_1, x_2 \hdots x_b \rbrace$ from the dataset and fix a hidden layer of the DNN, defining the map $f_i(x)$. Denote sizes of $x$, $f_i(x)$ by $n$ and $m$ correspondingly. Then, using the notation from \cref{sec:statement} we can compute $J_i(x_j) \in \mathbb{R}^{m \times n}$ for each $x_j \in X_b$. Let us now stack these Jacobian matrices vertically obtaining the matrix $J_i(X_b)$ of size $bm \times n$:
\begin{equation}\label{eq:big_jac}
J_i(X_b) = 
\begin{bmatrix}
J_i(x_1) \\
J_i(x_2) \\
\hdots \\
J_i(x_b)
\end{bmatrix}.
\end{equation}
Note that to compute the matvec functions of $J_i(X_b)$ and $J_i^{\top}(X_b)$ it suffices to be able to compute the individual matvec functions of $J_i(x)$ and $J_i^{\top}(x)$. We will present an algorithm for that in the next section and for now let us assume that these matvec functions are given. We can now apply \cref{alg:power} to the matrix $J_i(X_b)$ obtaining  Stochastic Power Method (SPM). 
\begin{algorithm}[!ht]
\caption{Stochastic Power Method for generating universal adversarial perturbations}\label{alg:uap}
\begin{algorithmic}[1]
\State \textbf{Inputs}: a batch of images $X_b = \lbrace x_1, x_2, \hdots x_b \rbrace$, $f_i(x)$ - fixed hidden layer of the DNN
\For{$x_j \in X$}
\State Construct the matvec functions of $J_i(x_j)$ and $J_i^{\top}(x_j)$  
\EndFor
\State Construct the matvec functions of $J_i(X_b)$ and $J_i^{\top}(X_b)$ defined in \cref{eq:big_jac} 
\State Run \cref{alg:power} with desired $p$ and $q$
\State \textbf{return} $\varepsilon$\Comment the universal perturbation
\end{algorithmic}
\end{algorithm}
Note that in \cref{alg:uap} we could in principle change the batch $X_b$ between iterations of the power method to compute "more general" singular vectors. However in our experiments we discovered that it almost does not affect the fooling rate of the generated universal perturbation.
\section{Efficient implementation of the matvec functions}
\label{sec:matvecs}
Matrices involved in \cref{alg:uap} for typical DNNs are too large to be formed explicitly. However, using automatic differentiation available in most deep learning packages it is possible to construct matvec functions which then are evaluated in a fraction of a second. To compute the matvecs we follow the well-known approach based on Pearlmutter's R-operator \cite{pearlmutter1994fast}, which could be briefly explained as follows. Suppose that we are given an operation $\grad_x[f](x_0)$ which computes the gradient of a scalar function $f(x)$ with respect to the vector variable $x$ at the point $x_0$. Let $f_i(x)$ be some fixed layer of the DNN, such that $x \in \mathbb{R}^n$ and $f_i(x) \in \mathbb{R}^m$, thus $J_i(x) \in \mathbb{R}^{m \times n}$ and for vectors $v_1 \in \mathbb{R}^n$, $v_2 \in \mathbb{R}^{m}$ we would like to compute $J_i(x) v_1$, $J_i^{\top}(x)v_2$ at some fixed point $x$. These steps are presented in \cref{alg:mbv}.
\\
For a given batch of images this algorithm is run only once.
\begin{algorithm}[!ht]
\caption{Constructing the matvec functions of the Jacobian matrix of a hidden layer of a DNN}\label{alg:mbv}
\begin{algorithmic}[1]
\State \textbf{Inputs} :$v_1 \in \mathbb{R}^n$, $v_2 \in \mathbb{R}^m$ - vectors to compute the matvec functions of, $f_i(x)$ - fixed hidden layer of the DNN
\State $J_i^{\top}(x)v_2 \gets \grad_v [\langle v, f_i(x) \rangle](v_2)$
\State $g(v_2) \gets \langle J_i^{\top}(x)v_2, v_1 \rangle$
\State $J_i(x) v_1 \gets \grad_{v_2} [g](\textbf{0}_m)$
\State \textbf{return} $J_i(x) v_1, J_i^{\top}(x)v_2$
\end{algorithmic}
\end{algorithm}

Let us summarize our approach for generating universal perturbations. Suppose that we have some dataset of natural images $X$ and a fixed deep neural network trained to perform image classification. At first we choose a fixed random batch of images $X_b$ from $X$ and specify a hidden layer of the DNN. Then using \cref{alg:mbv} we construct the matvec functions of the matrix defined by \cref{eq:big_jac}. Finally we run \cref{alg:uap} to obtain the perturbation and then rescale it if necessary.
\section{Experiments}
In this section we analyze various adversarial perturbations constructed as discussed in \cref{sec:matvecs}. For testing purposes we use the ILSVRC 2012 validation dataset \cite{russakovsky2015imagenet} ($50000$ images). 
\subsection{Adversarial perturbations}
In our experiments we chose $p=\infty, q=10$ and computed the $(p, q)$ - singular vectors for various layers of VGG-16 and VGG-19 \cite{simonyan2014very} and ResNet50 \cite{he2016deep}. $q=10$ was chosen to smoothen optimization problem and effectively serves as the replacement for $q = \infty$, for which the highest fooling rates were reported in \cite{moosavi2016universal}. We also investigate other values of $q$ in \cref{sec:analysis-q}. Batch size in \cref{alg:uap} was chosen to be $64$ and we used the same $64$ images to construct all the adversarial perturbations.
 \begin{figure*}[!htb]
 \centering
 \begin{subfigure}[!h]{0.65\textwidth}
  \includegraphics[width=1.0\textwidth]{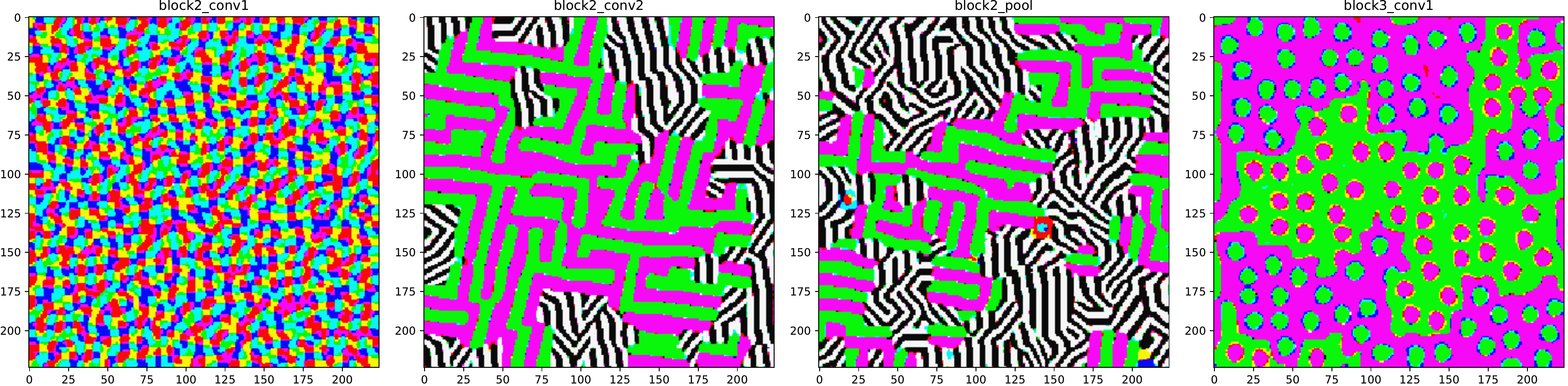}
 \caption{VGG-16}
 \label{fig:vgg16_singvect}
 \end{subfigure}
	\begin{subfigure}[!h]{0.65\textwidth}
  \includegraphics[width=1.0\textwidth]{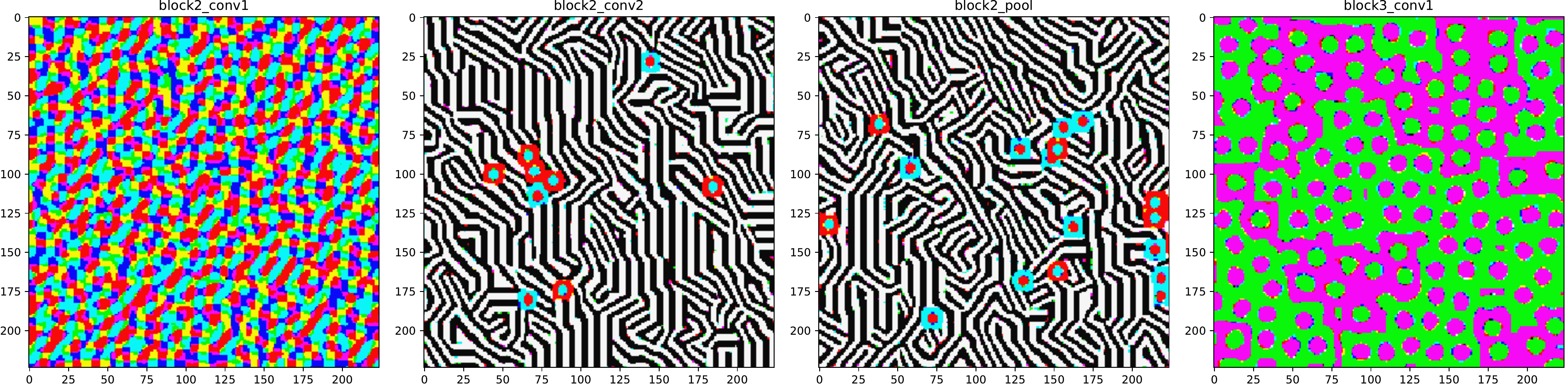}
 \caption{VGG-19}
 \label{fig:vgg19_singvect}
 \end{subfigure}
 \begin{subfigure}[!h]{0.65\textwidth}
  \includegraphics[width=1.0\textwidth]{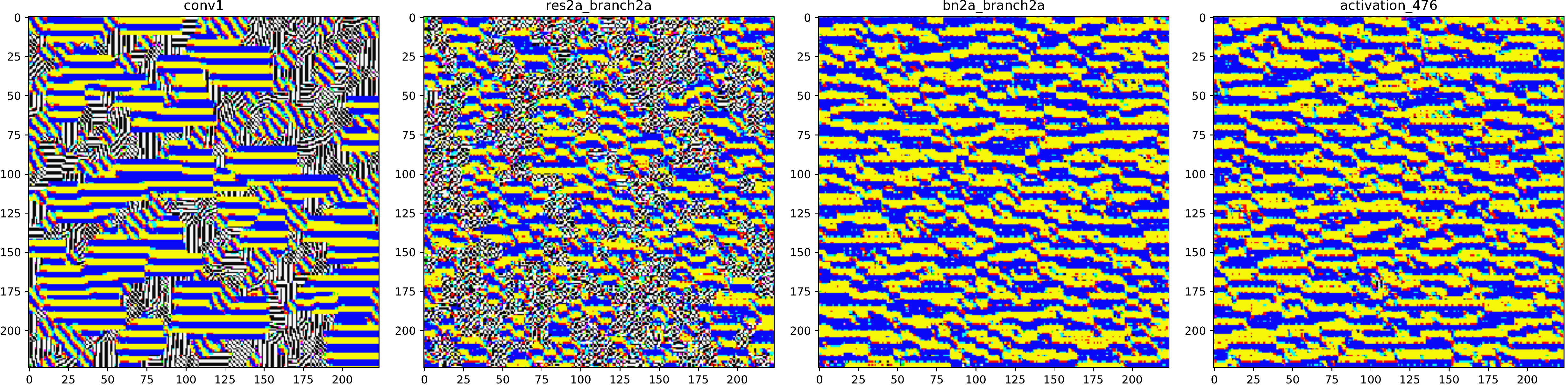}
 \caption{ResNet50}
 \label{fig:resnet_singvect}
 \end{subfigure}
 \caption{Universal adversarial perturbations constructed using various layers of various DNNs.}
 \label{fig:adv_ex}
 \end{figure*}
Some of the computed singular vectors are presented in \cref{fig:vgg16_singvect,fig:vgg19_singvect,fig:resnet_singvect}. We observe that computed singular vectors look visually appealing and present interesting visual patterns. Possible interpretation of these patterns can be given if we note that extremely similar images were computed in \cite{olah2017feature} in relation to \emph{feature visualization}. Namely, for various layers in GoogLeNet \cite{szegedy2015going} the images which activate a particular neuron were computed. In particular, visualization of the layer \texttt{conv2d0}, which corresponds to \emph{edge detection}, looks surprisingly similar to several of our adversarial perturbations. Informally speaking, this might indicate that adversarial perturbations constructed as the $(p, q)$-singular vectors attack a network by ruining a certain level of \emph{image understanding}, where in particular first layers correspond to edge detection. This is partly supported by the fact that the approach used for feature visualization in \cite{olah2017feature} is based on computing the Jacobian matrix of a hidden layer and maximizing the response of a fixed neuron, which is in spirit related to our method.
\\
To measure how strongly the $(p, q)$-singular vector disturbs the output of the hidden layer based on which it was constructed, we evaluate the corresponding singular value. We have computed it for all the layers of VGG-16, VGG-19 and ResNet50. Results are given in \cref{fig:sing_vals}. Note that in general singular values of the layers of ResNet50 are much smaller in magnitude than those of the VGG nets, which is further shown to roughly correspond to the obtained fooling rates.
\\
Convergence of \cref{alg:uap} is analyzed in \cref{fig:sing_vals_convergence}. We observe that a relatively low number of iterations is required to achieve good accuracy. In particular if each evaluation of the matvec functions takes $O(n)$ operations, the total complexity is $O(dn)$ for $d$ iterations, which for $d$ as small as $60$ is a big improvement compared to $O(n^2)$ of the exact algorithm.
\begin{figure*}[htb!]
    \centering
    \begin{subfigure}[h!]{0.30\textwidth}
        \centering
        \includegraphics[width=1.0\textwidth]{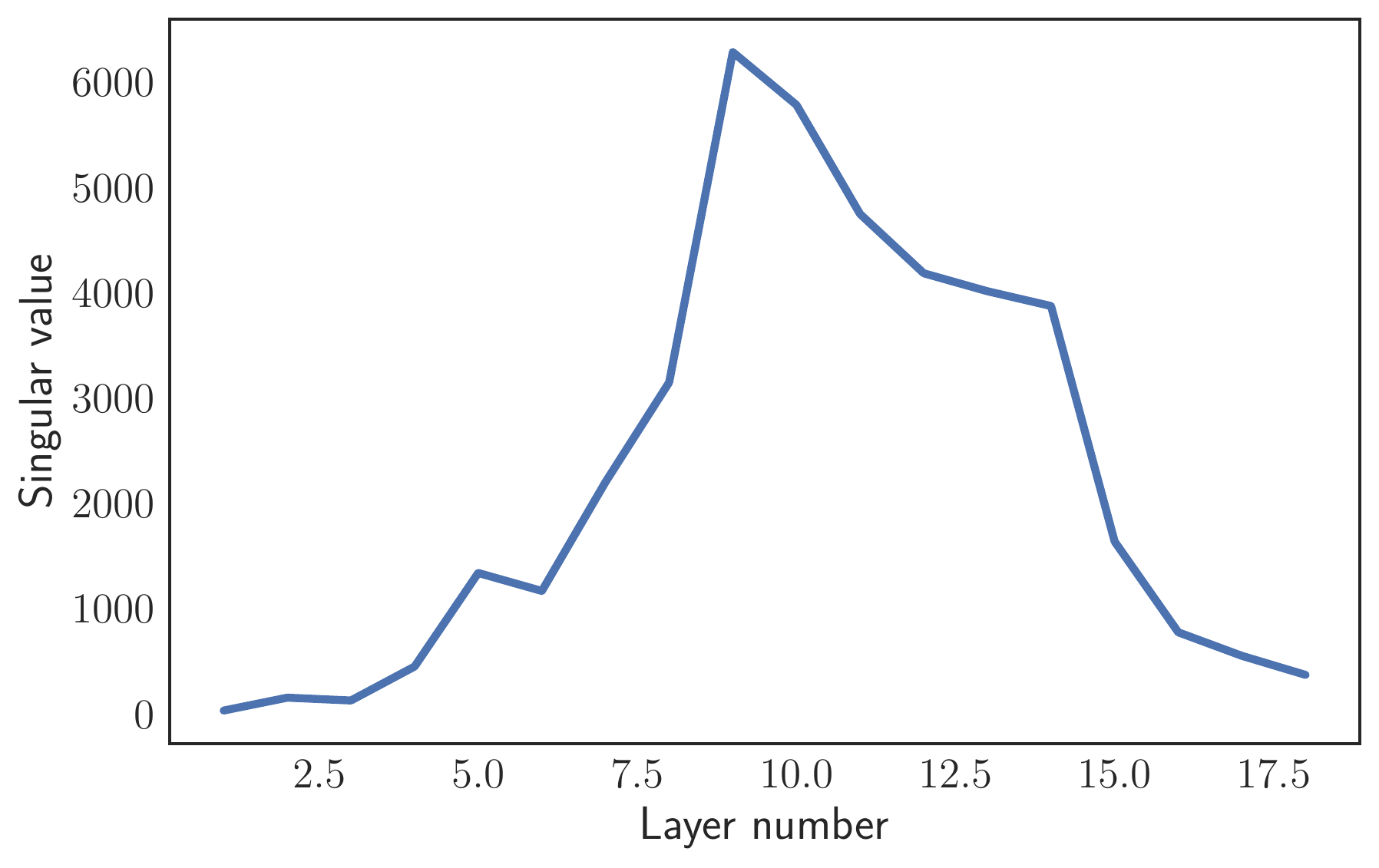}
        \caption{VGG-16}
    \end{subfigure}
    \begin{subfigure}[h]{0.30\textwidth}
        \centering
        \includegraphics[width=1.0\textwidth]{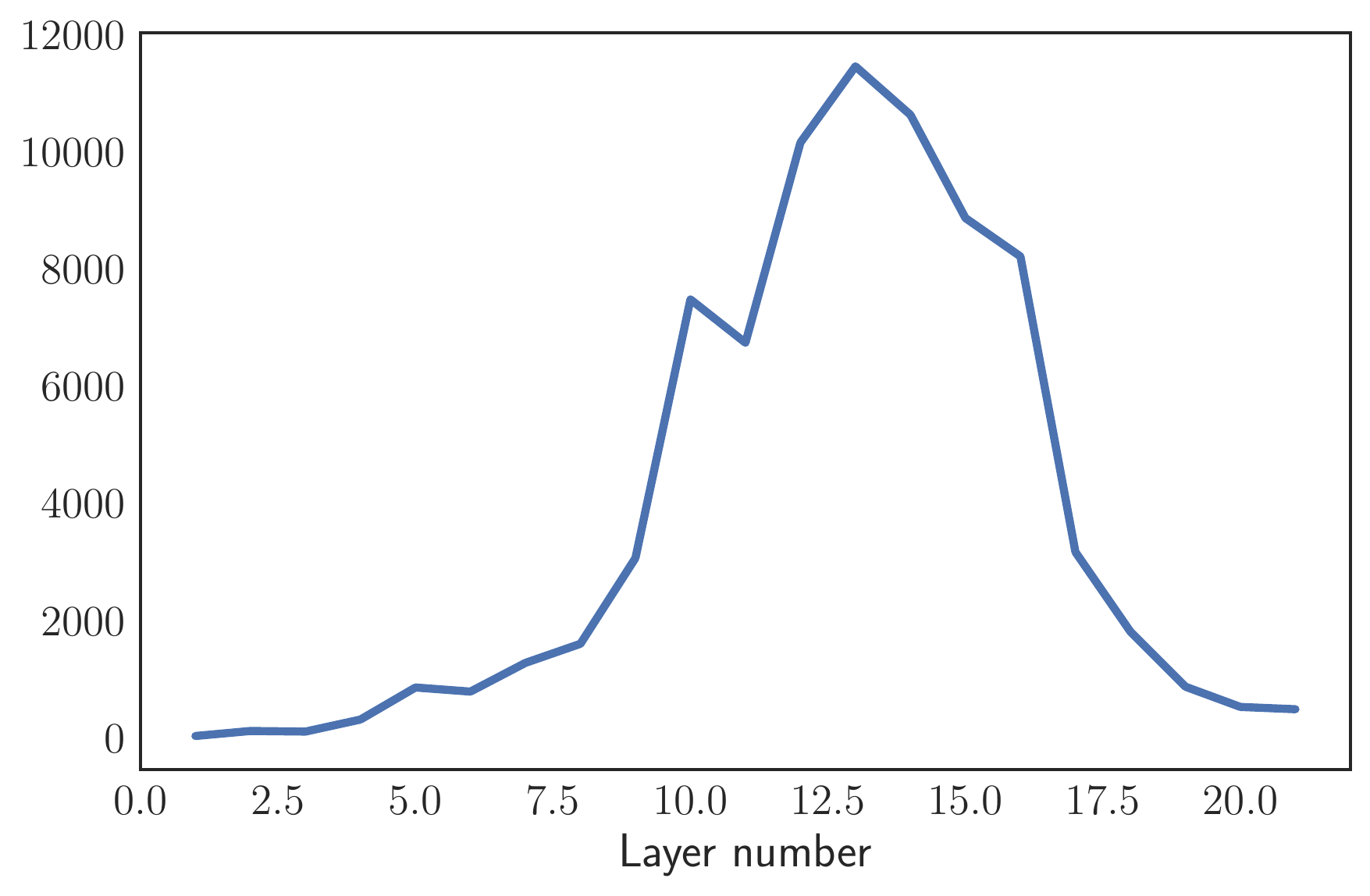}
        \caption{VGG-19}
    \end{subfigure}
        \begin{subfigure}[h]{0.30\textwidth}
        \centering
        \includegraphics[width=1.0\textwidth]{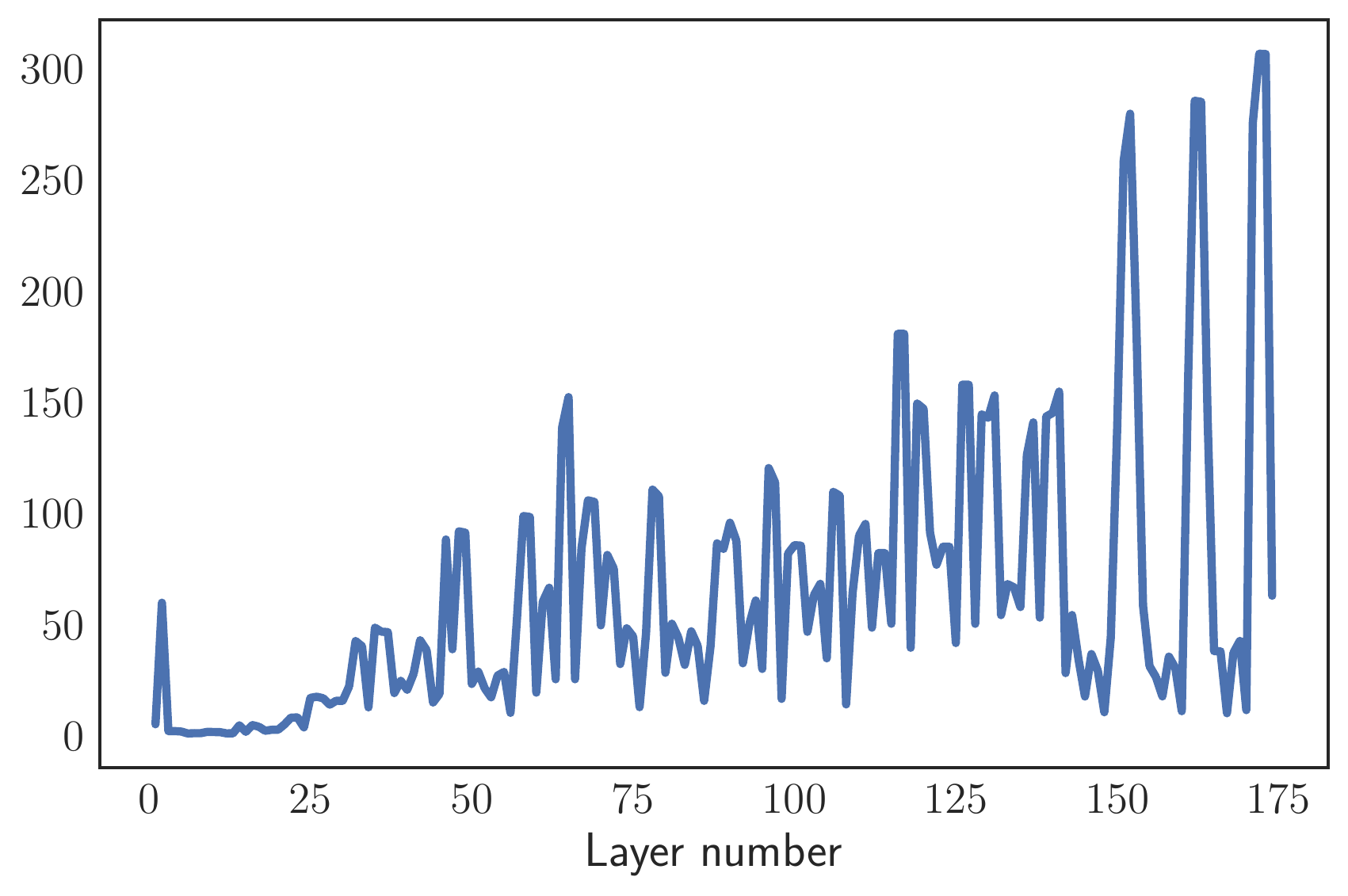}
        \caption{ResNet50}
    \end{subfigure}  
    \caption{$(p, q)$-singular values for all the layers of various DNNs. Values $p=\infty, q=10$ were used.}
    \label{fig:sing_vals}
\end{figure*}
\begin{figure}[htb]
\centering
\includegraphics[width=0.45\textwidth]{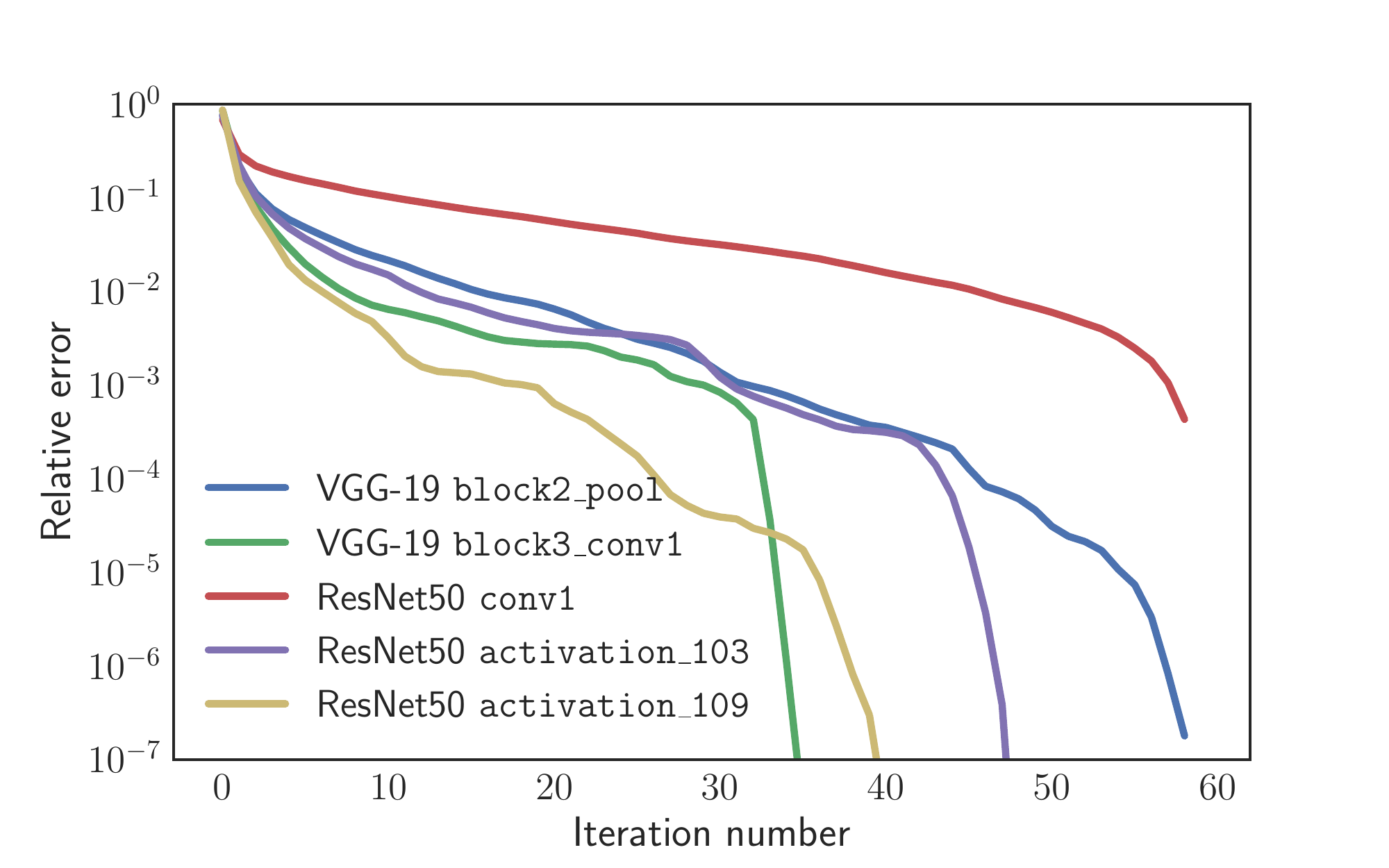}
\caption{Convergence of the $(\infty, 10)$-singular value. Relative error of the singular value w.r.t iteration number is shown.}
\label{fig:sing_vals_convergence}
\end{figure}

\subsection{Fooling rate and singular values}
\label{subsec:fr_sv}
\begin{figure}[htb]
 \centering
  \includegraphics[width=0.4\textwidth]{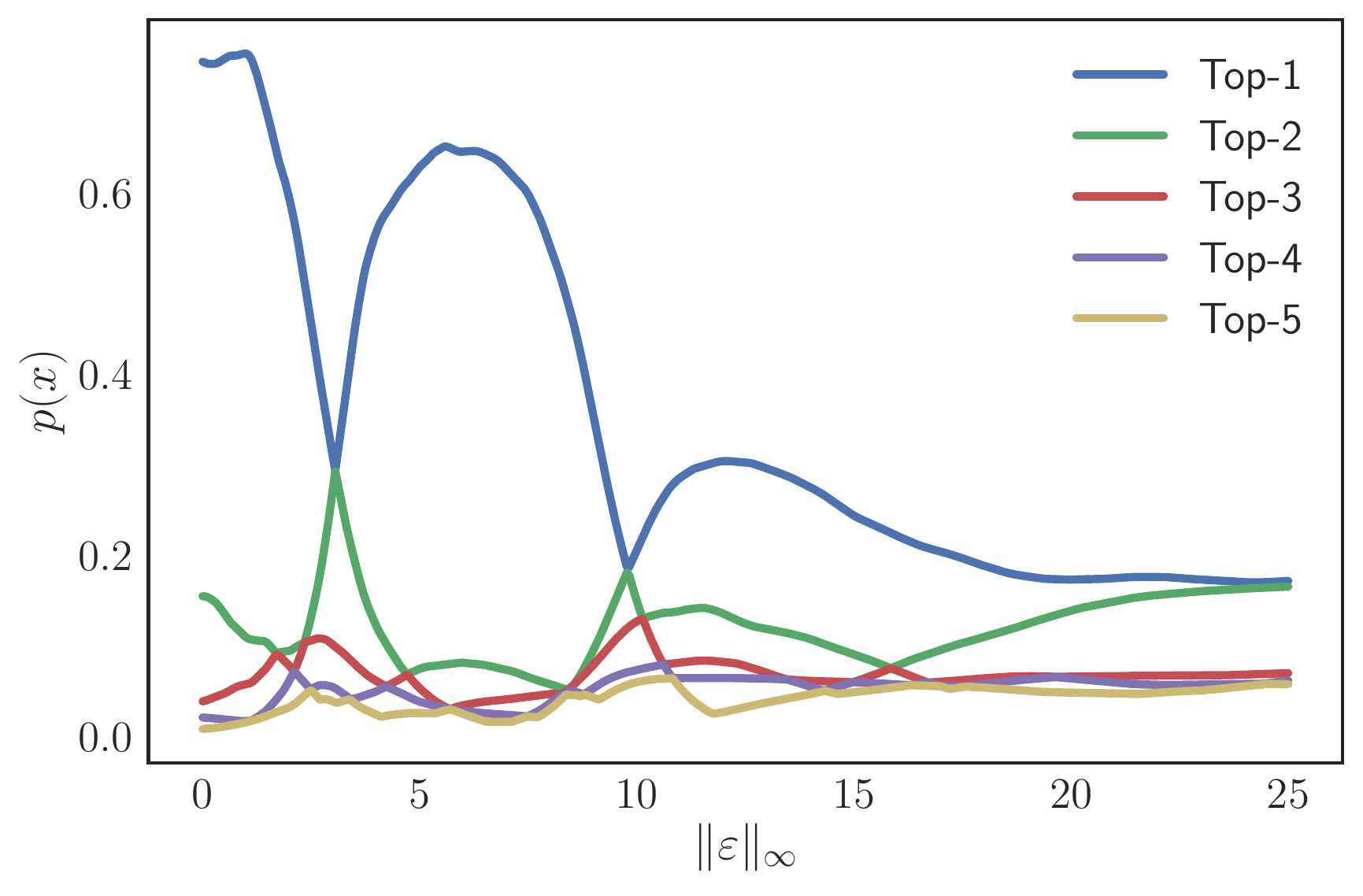}
 \caption{Top-5 probabilities predicted by the VGG-19 w.r.t $\infty$-norm of the universal adversarial perturbation. Tests were run for image 3 from \cref{fig:mis_images}. Universal adversarial perturbation with highest fooling rate in \cref{tab:vgg19} was chosen.}
 \label{fig:top5}
\end{figure}

\begin{figure}
\centering
\includegraphics[width=0.4\textwidth]{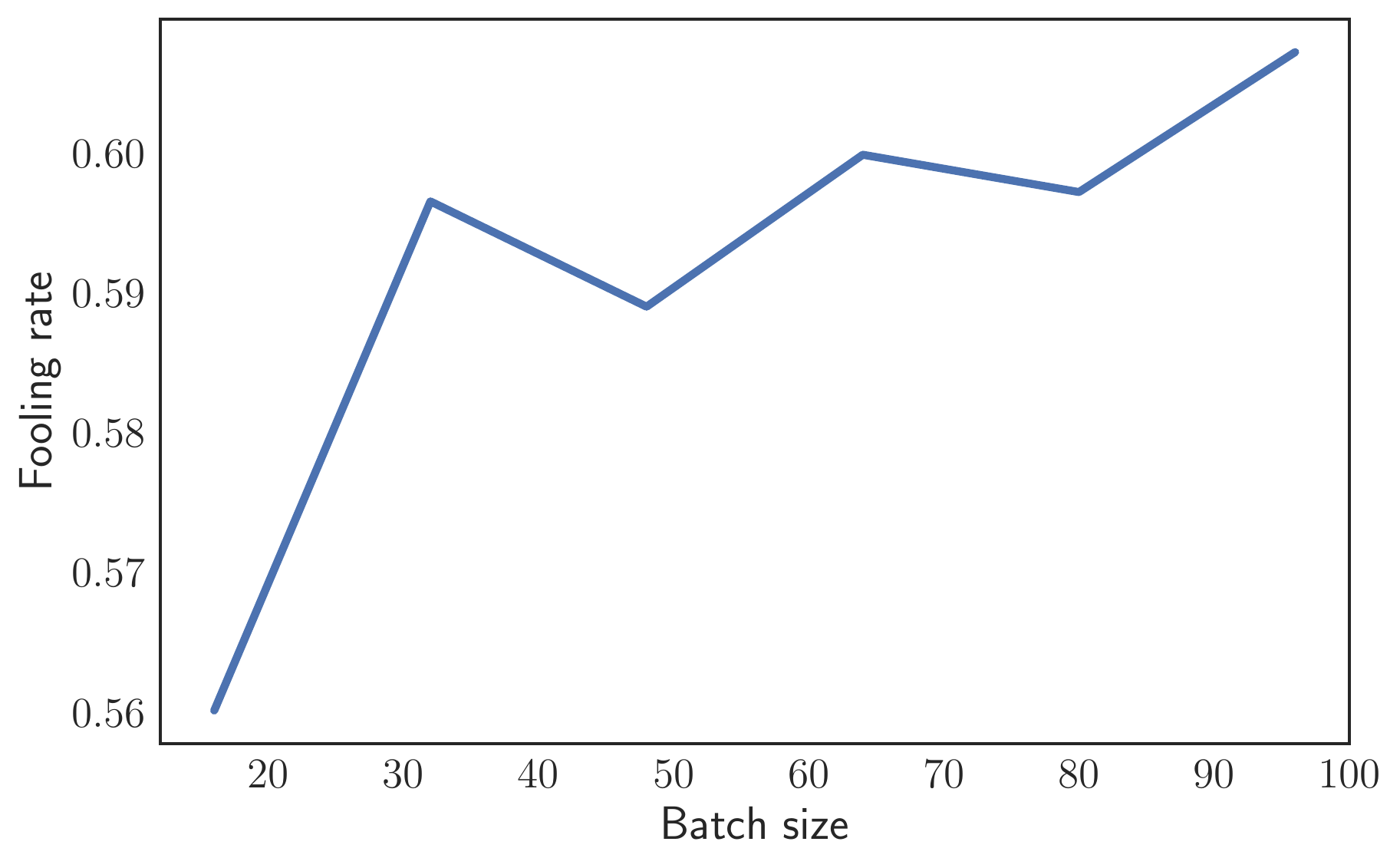}
\caption{Dependence of the fooling rate on the batch size. \texttt{block2\_pool} layer in VGG-19 was used for the experiment.}
\label{fig:fr_vs_batch}
\end{figure}

\begin{figure*}[htb]
 \centering
  \includegraphics[width=0.8\textwidth]{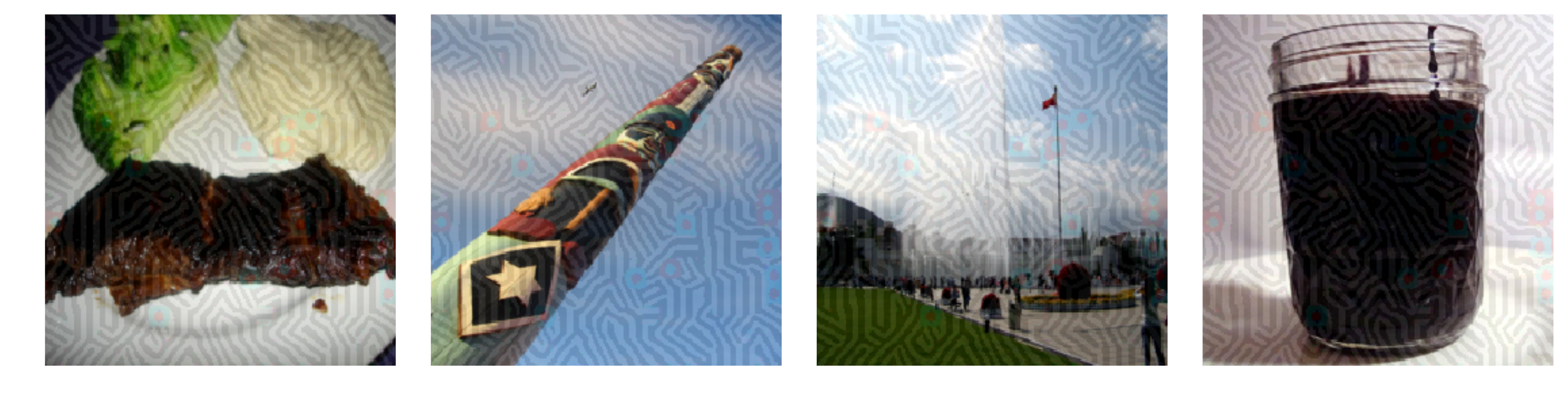}
 \caption{Examples of images misclassified after the adversarial attack (the attack based on \texttt{block2\_pool} layer of VGG-19 is shown). Predicted classes are given in \cref{tab:pred_mis_vgg16,tab:pred_mis_vgg19,tab:pred_mis_resnet50}.}
 \label{fig:mis_images} 
\end{figure*}

As a next experiment we computed and compared the fooling rate of the perturbation by various computed singular vectors. We choose \footnote{Pixels in the images from the dataset are normalized to be in $[0, 255]$ range, so by choosing $\|\varepsilon\|_{\infty} = 10$ we make the adversarial perturbations \textit{quasi-imperceptible} to human eye}$\|\varepsilon\|_{\infty} = 10$  -- recall that this can be achieved just by multiplying the computed singular vector by the factor of $10$. Results are given in \cref{tab:vgg16,tab:vgg19,tab:resnet}. 
\begin{table*}[htb!]
\centering
\begin{tabular}{lrrrr}
\toprule
{Layer name} &  block2\_pool &  block3\_conv1 &  block3\_conv2 &  block3\_conv3 \\
\midrule
\textbf{Singular value} &      1165.74 &       2200.08 &       3146.66 &       6282.64 \\
\textbf{Fooling rate  } &         \textbf{0.52} &          0.39 &          0.50 &          0.50 \\
\bottomrule
\end{tabular}
\caption{Fooling rates for VGG-16}
\label{tab:vgg16}
\centering
\begin{tabular}{lrrrr}
\toprule
{Layer name} &  block2\_pool &  block3\_conv1 &  block3\_conv2 &  block3\_conv3 \\
\midrule
\textbf{Singular value} &       784.82 &       1274.99 &       1600.77 &       3063.72 \\
\textbf{Fooling rate  } &         \textbf{0.60} &          0.33 &          0.50 &          0.52 \\
\bottomrule
\end{tabular}
\caption{Fooling rates for VGG-19}
\label{tab:vgg19}
\centering
\begin{tabular}{lrrrr}
\toprule
{Layer name} &  conv1 & res3c\_branch2a & bn5a\_branch2c & activation\_8 \\
\midrule
\textbf{Singular value} &         59.69 &         19.21 &  138.81 &           15.55 \\
\textbf{Fooling rate  } &          \textbf{0.44} &           0.35 &   0.34 &            0.34 \\
\bottomrule
\end{tabular}
\caption{Fooling rates for ResNet50}
\label{tab:resnet}
\end{table*}
\\
We see that using only $64$ images allowed us to achieve more than $40 \%$ fooling rate for all the investigated networks on the dataset containing $50000$ images of $1000$ different classes. This means that by analyzing less than $0.15 \%$ of the dataset it is possible to design strong universal adversarial attacks generalizing to many unseen classes and images. Similar fooling rates reported in \cite[Figure 6]{moosavi2016universal} required roughly $3000$ images to achieve (see \cref{sec:compare} for further comparison). Examples of images after addition of the adversarial perturbation with the highest fooling rate for VGG-19 are given in \cref{fig:mis_images}, and their predicted classes for various adversarial attacks (for each network we choose the adversarial perturbation with the highest fooling rate) are reported in \cref{tab:pred_mis_vgg16,tab:pred_mis_vgg19,tab:pred_mis_resnet50}. We note that the top-1 class probability for images after the adversarial attack is relatively low in most cases, which might indicate that images are moved away from the decision boundary. We test this behavior by computing the top-5 probabilities for several values of the $\infty$-norm of the adversarial perturbation. Results are given in \cref{fig:top5}. We see that top-$1$ probability decreases significantly and becomes roughly equal to the top-$2$ probability. Similar behavior was noticed in some of the cases when the adversarial example failed to fool the DNN --- top-$1$ probability still has decreased significantly. It is also interesting to note that such adversarial attack indeed introduces many new \emph{edges} in the image, which supports the claim made in the previous section.
\\
As a next experiment we investigate the dependence of the achieved fooling rate on the batch size used in \cref{alg:uap}. Some of the results are given in \cref{fig:fr_vs_batch}. Surprisingly, increasing the batch size does not significantly affect the fooling rate and by using as few as $16$ images it is possible to construct the adversarial perturbations with $56 \%$ fooling rate. This suggests that the singular vector constructed using Stochastic Power Method reasonably well approximates solution of the general optimization problem (\ref{eq:opt-big}). 
\\
It appears that higher singular value of the layer does not necessarily indicate higher fooling rate of the corresponding singular vector. However, as shown on \cref{fig:sing_vals} the singular values of various layers VGG-19 are in general larger than those of VGG-16, and of VGG-16 are in general larger than the singular values of ResNet50, which is roughly in correspondence between the maximal fooling rates we obtained for these networks. Moreover, layers closer to the input of the DNN seem to produce better adversarial perturbations, than those closer to the end. 
\\
Based on this observation we hypothesize that to defend the DNN against this kind of adversarial attack one can choose some subset $I$ of the layers (preferably closer to the input) of the DNN and include the term
$$\sum_{i \in I} \|J_i(X)\|_q^q,$$
in the regularizer, where $X$ indicates the current learning batch. We plan to analyze this approach in future work.
\\
Finally, we investigate if our adversarial perturbations generalize across different networks. For each DNN we have chosen the adversarial perturbation with the highest fooling rate from \cref{tab:vgg16,tab:vgg19,tab:resnet} and tested it against other networks. Results are given in \cref{tab:generalization}. We see that these adversarial perturbations are indeed \emph{doubly} universal, reasonably well generalizing to other architectures. Surprisingly, in some cases the fooling rate of the adversarial perturbation constructed using other network was \textit{higher} than that of the network's own adversarial perturbation. This universality might be explained by the fact that if Deep Neural Networks independently of specifics of their architecture indeed learn to detect low-level patterns such as edges, then adding an edge-like noise has a high chance to ruin the prediction. It is interesting to note that the adversarial perturbation obtained using \texttt{block2\_pool} layer of VGG-19 is the most efficient one, in correspondence with its interesting edge-like structure.
\begin{table}
\centering
\begin{tabular}{lrrr}
\toprule
{}  &  VGG-16 &  VGG-19 & ResNet50\\
\midrule
\textbf{VGG-16}   &    0.52 &    \textbf{0.60} & 0.39 \\
\textbf{VGG-19}   &    0.48 &    \textbf{0.60} & 0.38 \\
\textbf{ResNet50} &    0.41 &    \textbf{0.47} & 0.44 \\
\bottomrule
\end{tabular}
\caption{Generalization of the adversarial perturbations across networks. Columns indicate the DNN for which the adversarial perturbation was computed, rows indicate on which network it was tested. Adversarial perturbations with highest fooling rates in \cref{tab:vgg16,tab:vgg19,tab:resnet} were chosen.}
\label{tab:generalization}
\end{table}
\subsection{Dependence of the perturbation on $q$}
\label{sec:analysis-q}
\begin{figure*}[!htb]
 \centering
  \includegraphics[width=0.8\textwidth]{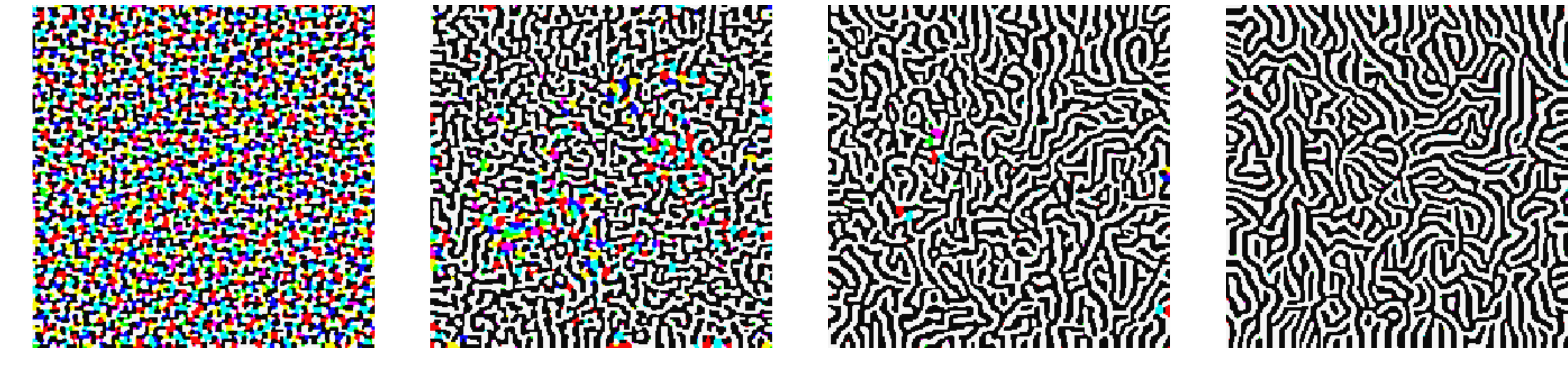}
 \caption{Adversarial perturbations constructed for various values of $q$. Presented images correspond to values $q$ uniformly increasing from $1.0$ to $5.0$. \texttt{block2\_pool} layer of VGG-19 was used.}
 \label{fig:adv_vs_norms}
\end{figure*}

\begin{figure}[!htb]
 \centering
  \includegraphics[width=0.4\textwidth]{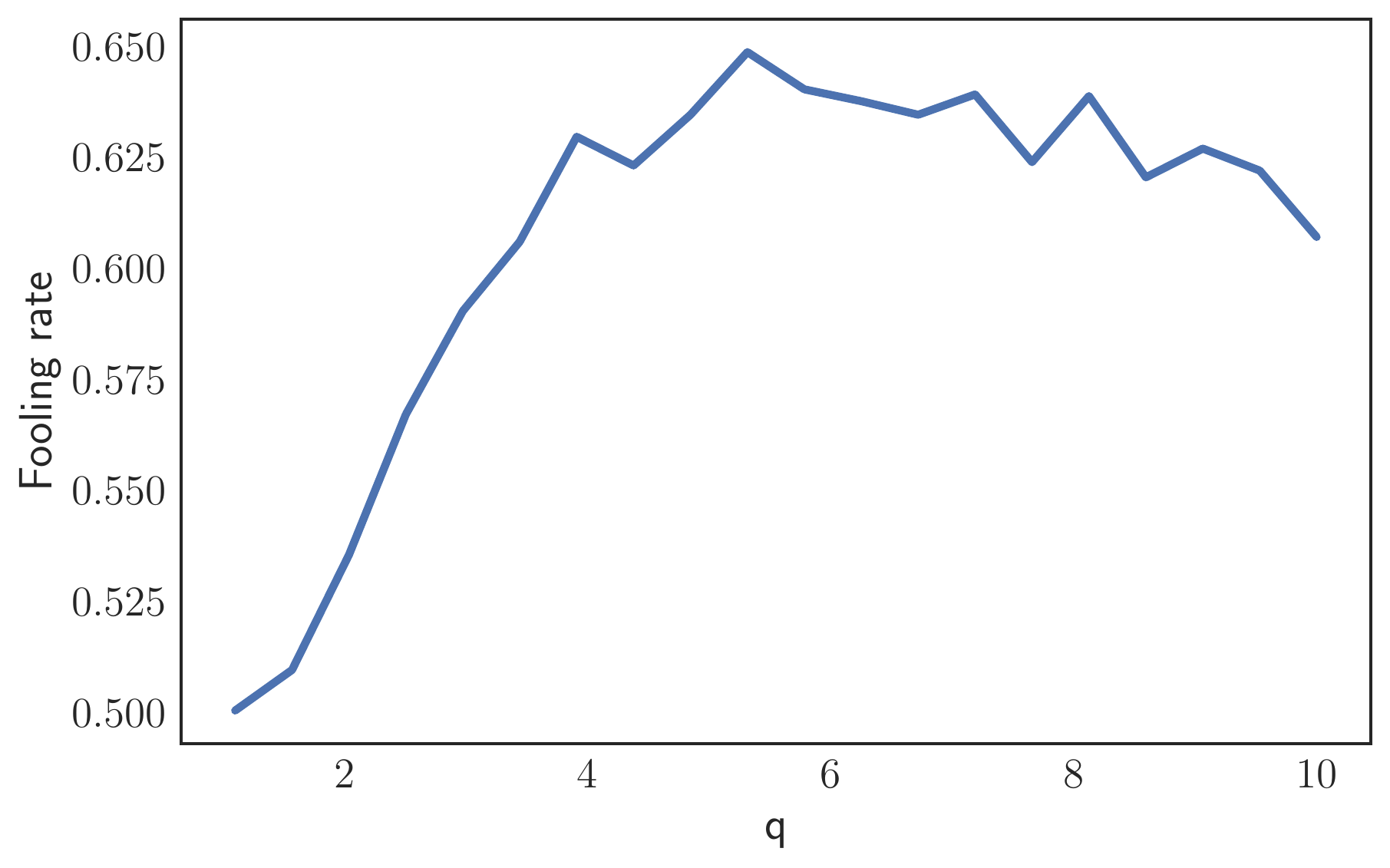}
 \caption{Dependence of the fooling rate on the value of $q$. As before, $p = \infty$ and norm of the adversarial perturbation was set to be $10$.}
 \label{fig:fr_vs_q}
\end{figure}

\begin{table*}[htb]
\centering
\begin{tabular}{lllll}
\toprule
{} & image\_1 & image\_2 & image\_3 & image\_4 \\
\midrule
$p(x)$   &  mashed\_potato $53.9\%$  & pole $37.6\%$ &  fountain $55.0\%$ & goblet $12.1\%$  \\
$p(x+\varepsilon)$ &  head\_cabbage $29.4 \%$  &rubber\_eraser $39.3\%$ & carousel $61.4 \%$ & bucket $36.6\%$\\
\bottomrule
\end{tabular}
\caption{VGG-16}
\label{tab:pred_mis_vgg16}
\centering
\begin{tabular}{lllll}
\toprule
{} & image\_1 & image\_2 & image\_3 & image\_4 \\
\midrule
$p(x)$   &  mashed\_potato $68.3\%$  & flagpole $37.4\%$ &  fountain $74.5\%$ & coffee\_mug $23.1\%$  \\
$p(x+\varepsilon)$ &  flatworm $26.5\%$  & letter\_opener $19.9\%$ & pillow $20.6\%$ & candle $40.3\%$ \\
\bottomrule
\end{tabular}
\caption{VGG-19}
\label{tab:pred_mis_vgg19}
\centering
\begin{tabular}{lllll}
\toprule
{} & image\_1 & image\_2 & image\_3 & image\_4 \\
\midrule
$p(x)$   & mashed\_potato $94.2\%$  & totem\_pole $43.1\%$ &  flagpole $35.3\%$ & chocolate\_sauce $22.0\%$  \\
$p(x+\varepsilon)$ &  stole $21.7\%$  &fountain\_pen $27.6\%$ & monitor $9.76\%$ & goblet $40.3\%$ \\
\bottomrule
\end{tabular}
\centering
\caption{ResNet50}
\label{tab:pred_mis_resnet50}
\end{table*}
In the analysis so far we have chosen $q=10$ as an approximate to $q = \infty$. However, any value of $q$ can be used for constructing the adversarial perturbations and in this subsection we investigate how the choice of $q$ affects the fooling rate and the generated perturbations (while keeping $p=\infty$). Perturbations computed for several different values of $q$ are presented in \cref{fig:adv_vs_norms}, and the corresponding fooling rates are reported in \cref{fig:fr_vs_q}. We observe that bigger values of $q$ produce more clear edge-like patterns, which is reflected in the increase of the fooling rate. However, the maximal fooling rate seems to be achieved at $q \approx 5$, probably because it is 'smoother' substitute for $q = \infty$ than $q = 10$, which might be important in such large scale problems.
\subsection{Comparison of the algorithms}\label{sec:compare}
\begin{figure}[htb]
 \centering
  \includegraphics[width=0.45\textwidth]{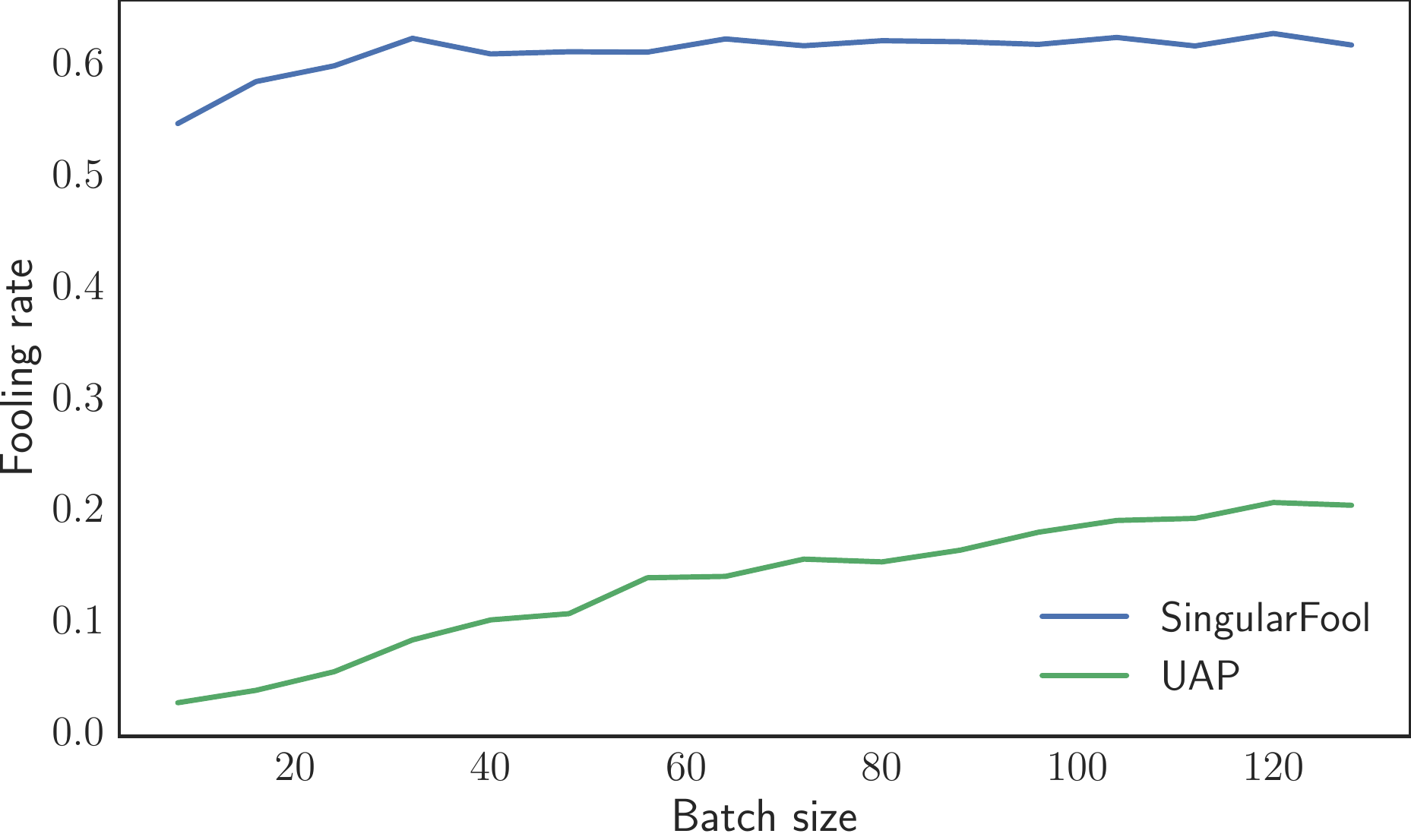}
 \caption{Dependence of the fooling rate on the number of images used for constructing the universal perturbation. SingularFool denotes the method proposed in the current paper, UAP denotes the algorithm presented in \cite{moosavi2016universal}. $q = 5$ and \texttt{block2\_pool} layer of VGG-19 were used.}
 \label{fig:compare}
\end{figure}

In this subsection we perform a comparison of the algorithm presented in Moosavi \etal, which we refer to as UAP, and our method. For the former we use the \texttt{Python} implementation \href{https://github.com/LTS4/universal/}{https://github.com/LTS4/universal/}. Since one of the main features of our method is an extremely low number of images used for constructing the perturbation, we decided to compare the fooling rates of universal perturbations constructed using these two methods for various batch sizes. Results are presented in \cref{fig:compare}. Note that our method indeed captures the universal attack vector relatively fast, and the fooling rate stabilizes on roughly $63\%$, while the fooling rate of the perturbation constructed by the UAP method starts low and then gradually increases as more images are added. Running time of our algorithm depends on which hidden layer we use. As an example for \texttt{block2\_pool} layer of VGG-19 the running time per iteration of the power method for a batch of one image was roughly $0.06$ seconds (one NVIDIA Tesla K80 GPU was used and the algorithm was implemented using \texttt{Tensorflow} \cite{tensorflow2015-whitepaper} and \texttt{numpy} libraries). Since the running time per iteration linearly depends on the batch size $b$, the total running time could be estimated as $0.06 b d$ seconds for $d$ iterations. By fixing $b=32$ and $d=30$ we obtain that the total running time to generate the universal perturbation with approximately $60\%$ fooling rate on the whole dataset is roughly $1$ minute (we did not include the time required to compute the symbolic Jacobian matvecs since it is performed only once, and is also required in the implementation of UAP, though different layer is used). In our hardware setup the running time of the UAP algorithm with batch size $128$ was approximately $10$ minutes, and the fooling rate of roughly $20 \%$ was achieved. According to \cite[Figure 6]{moosavi2016universal} approximately $3000$ images will be required to obtain the fooling rates of order $60 \%$. 
\section{Related work}
Many different methods \cite{goodfellow2014explaining, moosavi2016universal, kurakin2016adversarial, szegedy2013intriguing, liu2016delving} have been proposed to perform adversarial attacks on Deep Neural Networks in the \emph{white box} setting where the DNN is fully available to the attacker. Two works are especially relevant for the present paper. Goodfellow \etal \cite{goodfellow2014explaining} propose the fast gradient sign method, which is based on computing the gradient of the loss function $l(x)$ at some image $x$ and taking its $\sign$ as the adversarial perturbation. This approach allows one to construct rather efficient adversarial perturbations for individual images and can be seen as a particular case of our method. Indeed if we take the batch size to be equal to $1$ and the loss function $l(x)$ as the hidden layer, then $\sign \grad_{x} l(x)$ is exactly the solution of the problem (\ref{eq:max_problem}) with $p = 1$ and $L=1$ (since $l(x)$ is just a number this problem does not depend on $q$). Second work is Moosavi \etal \cite{moosavi2016universal} where the universal adversarial perturbations have been proposed. It is based on a sequential solution of nonlinear optimization problems followed by a projection onto $p = \infty$ ($p=2$) sphere, which iteratively computes the 'worst' possible direction towards the decision boundary. Optimization problems proposed in the current work are simpler in nature and well-studied, and due to their homogeneous property the adversarial perturbation with an arbitrary norm is obtained by simply rescaling the once computed perturbation, in contrast with the algorithm in \cite{moosavi2016universal}. 
\section{Conclusion}
In this work we explored a new algorithm for generating universal adversarial perturbations and analyzed their main properties, such as generalization across networks, dependence of the fooling rate on various hyperparameters and having certain visual properties. We have showed that by using only $64$ images a single perturbation fooling the network in roughly $60 \%$  cases can be constructed, while the previous known approach required several thousand of images to obtain such fooling rates. In a future work we plan to address the relation between feature visualization \cite{olah2017feature} and adversarial perturbations, as well as analyzing the defense approach discussed in \cref{subsec:fr_sv}.
\section*{Acknowledgements}
This study was supported by the Ministry of Education and
Science of the Russian Federation (grant 14.756.31.0001),
by RFBR grants 16-31-60095-mol-a-dk, 16-31-00372-mola
and by Skoltech NGP program.
{\small
\bibliographystyle{ieee}
\bibliography{power}
}
\end{document}